\documentclass[11pt]{article}
\usepackage[utf8x]{inputenc}
\usepackage[hebrew,english]{babel}
\usepackage{times}
\usepackage{url}
\usepackage{latexsym}
\usepackage[usenames, dvipsnames]{color}
\usepackage{eacl2017}

\newcommand{\heb}[1]{\begin{small}\R{#1}\end{small}}

\definecolor{correct}{RGB}{26, 134, 58}
\definecolor{wrong}{RGB}{255, 0, 0}

\eaclfinalcopy 

\title{The Interplay of Semantics and Morphology in Word Embeddings}

\author{Oded Avraham \and Yoav Goldberg \\
		Computer Science Department \\
	    Bar-Ilan University\\       
	    Ramat-Gan, Israel\\
	    {\tt \{oavraham1,yoav.goldberg\}@gmail.com}}

\date{}

\begin{document}
\maketitle
\begin{abstract}
We explore the ability of word embeddings to capture both semantic and morphological similarity, as affected by the different types of linguistic properties (surface form, lemma, morphological tag) used to compose the representation of each word. We train several models, where each uses a different subset of these properties to compose its representations. By evaluating the models on semantic and morphological measures, we reveal some useful insights on the relationship between semantics and morphology.
\end{abstract}

\section{Introduction}
Word embedding models learn a space of continuous word representations, in which similar words are expected to be close to each other. Traditionally, the term \textit{similar} refers to \textit{semantic} similarity (e.g. \textit{walking} should be close to \textit{hiking}, and \textit{happiness} to \textit{joy}), hence the model performance is usually evaluated using semantic similarity datasets. Recently, several works introduced morphology-driven models  
motivated by the
poor performance of traditional models on morphologically complex words. Such words are often rare, and there is not enough evidence to model them correctly. The morphology-driven models allow pooling evidence from different words which have the same base form. These models work by learning per-morpheme representations rather than just per-word ones, and compose the representing vector of each word from those of its morphemes -- as derived from a supervised or unsupervised morphological analysis -- and (optionally) its surface form (e.g. \textit{walking} = $f(v_{walk}, v_{ing}, v_{walking})$). 

The works differ in the way they acquire morphological knowledge (from using
linguistically derived morphological analyzers on one end, to approximating
morphology using substrings while relying on the concatenative nature of
morphology, on the other) and in the model form (cDSMs
\cite{lazaridou2013compositional}, RNN \cite{luong2013better}, LBL
\cite{botha2014compositional}, CBOW \cite{qiu2014co}, SkipGram
\cite{soricut2015unsupervised,bojanowski2016enriching}, GGM \cite{cotterell2016morphological}). But essentially, they
all show that breaking a word into morphological components (base form,
affixes and potentially also the complete surface form), learning a vector
for each component, and representing a word as a composition of these vectors
 improves the models semantic performance, especially on rare words.\smallskip \\
\noindent\textbf{In this work} we argue that these models capture two distinct aspects of word similarity,
\emph{semantic} (e.g. \textit{sim(walking, hiking)} $>$
\textit{sim(walking, eating)}) and \emph{morphological} (e.g. 
\textit{sim(walking, hiking)} $>$ \textit{sim(walking, hiked)}), and that these
two aspects are at odds with each other (should \textit{sim(walking, hiking)} be
lower or higher than \textit{sim(walking, walked)}?).
The \textit{base form} component of the compositional models is mostly
responsible for semantic aspects of the similarity, while the
\textit{affixes} are mostly responsible for morphological similarity.

This analysis brings about several natural questions: 
is the combination of semantic and morphological components used in previous
work ideal for every purpose? For example, if we exclude the morphological component
from the representations, wouldn't it improve the semantic performance?
What is the contribution of using the surface form? And do the models behave
differently on common and rare words?
We explore these questions in order to help the users of
morphology-driven models choose the right configuration for their
needs: semantic or morphological performance, on common or rare words.

We compare different configurations of morphology-driven models, while controlling for
the components composing the representation.
We then separately evaluate the semantic and morphological performance of each
model, on rare and on common words.
We focus on \textit{inflectional} (rather than \textit{derivational}) morphology. This is due to the fact that derivations (e.g. \textit{affected} $\rightarrow$  \textit{unaffected}) often drastically change the meaning of the word, and therefore the benefit of having similar representations for words with the same derivational base is questionable, as discussed by Lazaridou et al~\shortcite{lazaridou2013compositional} and Luong et al~\shortcite{luong2013better}. Inflections (e.g. \textit{walked} $\rightarrow$ \textit{walking}), in contrast, preserve the word lexical meaning, and only change its grammatical categories values.

Our experiments are performed on Modern Hebrew, a language with rich
inflectional morphological system. 
We build on a recently introduced evaluation dataset
for semantic similarity in Modern Hebrew \cite{avraham2016improving}, which we further extend with a collection of rare words. We also create datasets for morphological
similarity, for common and rare words.  
Hebrew's morphology is not concatenative, so unlike
most previous work we do not break the words into base and affixes, but instead
rely on a morphological analyzer and represent words using their \emph{lemmas}
(corresponding to the base form) and their \emph{morphological tags} (from which the
morphological forms are derived, corresponding to affixes). This allow us
to have a finer grained control over the composition, separating inflectional
from derivational processes. We also compare to a strong character ngram based
model, that mixes the different components and does not allow finer-grained
distinctions.

We observe a clear trade-off between the morphological and semantic performance -- models that excel on one metric perform badly on the other.
We present the strengths and weaknesses of the different configurations, to help
the users choose the one that best fits their needs. 

We believe that this work is the first to make a comprehensive
comparison between various configurations of morphology-driven models:
among the previous work mentioned above, only few explored configurations other
than (base + affixes) or (surface + base + affixes). Lazaridou et
al~\shortcite{lazaridou2013compositional} and Luong et
al~\shortcite{luong2013better} trained models which represent a word by its base
only, and showed that these models performs worse than the compositional ones
(base + affixes). However, the poor results for the base-only models were mainly
attributed to undesirable capturing of derivational similarity, e.g.
\textit{(affected, unaffected)}. Working with a more linguistically informed
morphological analyzer allows us to tease apart inflectional from derivational
processes, leading to different results.

Most of the works on morphology-driven models evaluate the semantic performance of the models, while others perform morphological evaluation. To the best of our knowledge, this work is the first to evaluate both aspects. While our experiments focus on Modern Hebrew due to the availability of a reliable semantic similarity dataset, we believe our conclusions hold more generally.


\section{Models} 
Our model form is a generalization of the fastText model \cite{bojanowski2016enriching}, which in turn extends the skip-gram model of Mikolov et al~\shortcite{mikolov2013efficient}.
The skip-gram model takes a sequence of words $w_1,...,w_T$ and a function $s$
assigning scores to (word, context) pairs, and maximizes
\[\sum_{t=1}^{T}\left(\sum_{w_c\in \mathcal{C}_t}\ell(s(w_t,w_c))+\sum_{w'_c\in
\mathcal{N}_{t}}\ell(-s(w_t,w'_c))\right)\] where $\ell$ is the log-sigmoid loss
function, $\mathcal{C}_t$ is a set of
context words, and $\mathcal{N}_{t}$ is a set of negative examples sampled
from the vocabulary. $s(w_t,w_c)$ is defined as
$s(w_t,w_c)=\mathbf{v}_{w_t}^{\top} \mathbf{u}_{w_c}$ (where $\mathbf{v}_{w_t}$
and $\mathbf{u}_{w_c}$ are the embeddings of the focus and the context words).

Bojanowski et al~\shortcite{bojanowski2016enriching} replace the word
representation $\mathbf{v}_{w_t}$ with the set of character ngrams appearing in it:
$\mathbf{v}_{w_t}=\sum_{g\in\mathcal{G}(w_t)}\mathbf{v}_{g}$ where
$\mathcal{G}(w_t)$ is the set of n-grams appearing in $w_t$. The n-grams are
used to approximate the morphemes in the target word.


We generalize Bojanowski et al~\shortcite{bojanowski2016enriching} by replacing
the set of ngrams $\mathcal{G}(w)$ with a set $\mathcal{P}(w)$ of explicit
linguistic properties. Each word $w_t$ is then composed as the sum of the
vectors of its
linguistic properties: $\mathbf{v}_{w_t} =
\sum_{p\in\mathcal{P}(w_t)}\mathbf{v}_p$. The linguistic properties we consider are the surface form of the word (W), it's
lemma (L) and its morphological tag (M)\footnote{The lemma and morphological tag for a word in context are obtained using a morphological analyzer and disambiguator. Then, each value of lemma/tag/surface from is associated with a trainable embedding vector.}. The lemma corresponds to the
base-form, and the morphological tag encodes the
grammatical properties of the word, from which its inflectional affixes are
derived (a similar approach was taken by Cotterell and
Sch{\"u}tze~\shortcite{cotterell2015morphological}).
Moving from a set of n-grams to a
set of explicit linguistic properties, allows finer control of the kinds
of information in the word representation. We train models with different
subsets of $\{W,L,M\}$.


\section{Experiments and Results}
Our implementation is based on the
\textit{fastText}\footnote{https://github.com/facebookresearch/fastText} library
\cite{bojanowski2016enriching}, which we modify as described above.
We train the models on the Hebrew Wikipedia ($\sim$4M sentences), using a window
size of 2 to each side of the focus word, and dimensionality of 200.
 We use the morphological disambiguator of
Adler~\shortcite{adler2007hebrew} to assign words with their morphological tags,
and the inflection dictionary of \textit{MILA}~\cite{mila2008} to find their
lemmas. For example, for the words \heb{נסתכל} (\textit{[we will] look [at]}),
\heb{הסתכלה} (\textit{[she] looked [at]}) and \heb{הסתכל} (\textit{[he] looked
[at]}) are assigned the tags \textit{VB.MF.P.1.FUTURE}, \textit{VB.F.S.3.PAST}
and \textit{VB.M.S.3.PAST} respectively, and share the lemma \heb{הסתכל}. We
train the models for the subsets $\{W\}$, $\{L\}$, $\{W,L\}$, $\{W,M\}$ and
$\{W,L,M\}$, as well as the original fastText (n-grams) model. Finally, we evaluate each model on several datasets, using both semantic and morphological performance measures.\footnote{Our code is available on \url{https://github.com/oavraham1/prop2vec}, our datasets on \url{https://github.com/oavraham1/ag-evaluation}}

\begin{table*}
\small
\begin{center}
\begin{tabular}{c|l|l|l}
	
     & 1st & 2nd & 3rd \\\hline
    \textit{W} & \heb{הביטה}:\textcolor{correct}{gaze}:\textcolor{correct}{VB}.\textcolor{correct}{F}.\textcolor{correct}{S}.\textcolor{correct}{3}.\textcolor{correct}{PAST} & \heb{חייכה}:\textcolor{wrong}{smile}:\textcolor{correct}{VB}.\textcolor{correct}{F}.\textcolor{correct}{S}.\textcolor{correct}{3}.\textcolor{correct}{PAST} & \heb{מתייפחת}:\textcolor{wrong}{cry}:\textcolor{correct}{VB}.\textcolor{correct}{F}.\textcolor{correct}{S}.\textcolor{correct}{3}.\textcolor{wrong}{PRESENT}\\\hline
    \textit{L} & \heb{הביטי}:\textcolor{correct}{gaze}:\textcolor{correct}{VB}.\textcolor{correct}{F}.\textcolor{correct}{S}.\textcolor{wrong}{2}.\textcolor{wrong}{IMPERATIVE} & \heb{התבונן}:\textcolor{correct}{watch}:\textcolor{correct}{VB}.\textcolor{wrong}{M}.\textcolor{correct}{S}.\textcolor{correct}{3}.\textcolor{correct}{PAST} & \heb{בהו}:\textcolor{correct}{stare}:\textcolor{correct}{VB}.\textcolor{correct}{MF}.\textcolor{wrong}{P}.\textcolor{correct}{3}.\textcolor{correct}{PAST} \\\hline
    \textit{WL} & \heb{נביט}:\textcolor{correct}{gaze}:\textcolor{correct}{VB}.\textcolor{correct}{MF}.\textcolor{wrong}{P}.\textcolor{wrong}{1}.\textcolor{wrong}{FUTURE} & \heb{התבוננה}:\textcolor{correct}{watch}:\textcolor{correct}{VB}.\textcolor{correct}{F}.\textcolor{correct}{S}.\textcolor{correct}{3}.\textcolor{correct}{PAST} & \heb{בוהה}:\textcolor{correct}{stare}:\textcolor{correct}{VB}.\textcolor{correct}{F}.\textcolor{correct}{S}.\textcolor{correct}{3}.\textcolor{wrong}{PRESENT} \\\hline
    \textit{WM} & \heb{חייכה}:\textcolor{wrong}{smile}:\textcolor{correct}{VB}.\textcolor{correct}{F}.\textcolor{correct}{S}.\textcolor{correct}{3}.\textcolor{correct}{PAST} & \heb{נחבלה}:\textcolor{wrong}{injure}:\textcolor{correct}{VB}.\textcolor{correct}{F}.\textcolor{correct}{S}.\textcolor{correct}{3}.\textcolor{correct}{PAST} & \heb{נשפה}:\textcolor{wrong}{blow}:\textcolor{correct}{VB}.\textcolor{correct}{F}.\textcolor{correct}{S}.\textcolor{correct}{3}.\textcolor{correct}{PAST} \\\hline
     \textit{LM} & \heb{הביטה}:\textcolor{correct}{gaze}:\textcolor{correct}{VB}.\textcolor{correct}{F}.\textcolor{correct}{S}.\textcolor{correct}{3}.\textcolor{correct}{PAST} & \heb{התבוננה}:\textcolor{correct}{watch}:\textcolor{correct}{VB}.\textcolor{correct}{F}.\textcolor{correct}{S}.\textcolor{correct}{3}.\textcolor{correct}{PAST} & \heb{זזה}:\textcolor{wrong}{move}:\textcolor{correct}{VB}.\textcolor{correct}{F}.\textcolor{correct}{S}.\textcolor{correct}{3}.\textcolor{correct}{PAST} \\\hline
     \textit{WLM} & \heb{הביטה}:\textcolor{correct}{gaze}:\textcolor{correct}{VB}.\textcolor{correct}{F}.\textcolor{correct}{S}.\textcolor{correct}{3}.\textcolor{correct}{PAST} & \heb{התבוננה}:\textcolor{correct}{watch}:\textcolor{correct}{VB}.\textcolor{correct}{F}.\textcolor{correct}{S}.\textcolor{correct}{3}.\textcolor{correct}{PAST} & \heb{פסעה}:\textcolor{wrong}{walk}:\textcolor{correct}{VB}.\textcolor{correct}{F}.\textcolor{correct}{S}.\textcolor{correct}{3}.\textcolor{correct}{PAST} \\\hline
\end{tabular}
\end{center}
\caption{\label{qualitative} Top-3 similarities for the word
\heb{הסתכלה} (\textit{[she] looked [at]}).\newline \footnotesize Each entry is of the form
\textit{ [word:lexical meaning:morphological tag]}. Green-colored items share
the semantic/inflection of the target word, while red-colored indicate a
divergence. In the morphological tags: M/F/MF indicate masculine/feminine/both, P/S indicate plural/singular, 1/2/3 indicate 1st/2nd/3rd person.}
\end{table*}

\paragraph{Semantic Evaluation Measure}
The common datasets for semantic similarity\footnote{E.g., WordSim353
\cite{finkelstein2001placing}, RW \cite{luong2013better} and SimLex999
\cite{hill2015simlex}} have some notable
shortcomings as noted in
\cite{avraham2016improving,faruqui2016problems,batchkarov2016critique,linzen2016issues}.
We use the evaluation method (and corresponding Hebrew similarity dataset) that we have introduced in a previous work ~\cite{avraham2016improving} (AG). The AG method
defines an annotation task which is more natural for human judges, resulting in
datasets with improved annotator-agreement scores. Furthermore, the AG's evaluation metric takes annotator agreement into account, by putting less weight on similarities that have lower annotator agreement.

An AG dataset is a collection of target-groups, where each group contains a target word (e.g. \emph{singer}) and three types of
candidate words:
\textit{positives} which are words ``similar'' to the target (e.g.
\emph{musician}),
\textit{distractors} which are words ``related but dissimilar'' to the target
(e.g. \emph{microphone}), and \textit{randoms} which are not related to the
target at all (e.g \emph{laptop}). The human annotators are asked to rank the
positive words by their similarity to the target word (distractor and random
words are not annotated by humans and are automatically ranked below the
positive words). This results in a set of triples of
a target word $w$ and two candidate words $c_1,c_2$, coupled with a
value indicating the confidence of ranking $sim(w,c_1)>sim(w,c_2)$ by the
annotators. A model is then scored based on its ability to correctly rank each
triple, giving more weight to highly-confident triples. The scores range between
0 (all wrong answers) to 1 (perfect match with human annotators).

We use this method on two datasets: the AG dataset from \cite{avraham2016improving}
(\textit{SemanticSim}, containing 1819 triples), and a new dataset we created in order to evaluate the models on rare words (similar to RW \cite{luong2013better}).
The rare-words dataset (\textit{SemanticSimRare}) follows the structure of \textit{SemanticSim}, but
includes only target words that occur less than 100 times in the corpus. It
contains a total of 163 triples, all of the type positive vs. random (we find
that for rare words, distinguishing similar words from random ones is a hard enough task for the models).

\paragraph{Morphological Evaluation Measure} 
Cotterrel and Sch\"utze \shortcite{cotterell2015morphological} introduced the
MorphoDist$_k$ measure, which quantifies the amount of morphological difference
between a target word and a list of its $k$ most similar words. 
We modify MorphoDist$_k$ measure to derive MorphoSim$_k$, a measure
that ranges between 0 and 1, where 1 indicates total morphological
compatibility.
The MorphoDist measure is defined as:
$MorphoDist_k(w) = \sum_{w'\in
\mathcal{K}_w}\min_{m_w,m_{w'}}d_h(m_w,m_{w'})$ where $\mathcal{K}_w$ is the
set of top-$k$ similarities of $w$, $m_w$ and $m_{w'}$ are possible
morphological tags of $w$ and $w'$ respectively (there may be more than one
possible morphological interpretation per word), and $d_h$ is the Hamming
distance between the morphological tags. \textit{MorphoDist} counts the \textit{total number}
of incompatible morphological components.
\textit{MorphoSim} calculates the \textit{average rate} of
\textit{compatible} morphological values. More formally,
$MorphoSim_k(w)=1-\frac{MorphoDist_k(w)}{k\cdot |m_w|}$, where $|m_w|$ is the number
of grammatical components specified in $w$'s morphological tag.

We use $k$=10 and calculate the average \textit{MorphoSim} score over 100 randomly chosen words. To evaluate the morphological performance on rare words, we run another benchmark (\textit{MorphoSimRare}) in which we calculate the average \textit{MorphoSim} score over the 35 target words of the \textit{SemanticSimRare} dataset.

\paragraph{Qualitative Results}
To get an impression of the differences in behavior between the models, we
queried each model for the top similarities of several words (calculated by
cosine similarity between words vectors), focusing on rare words. Table
\ref{qualitative} presents the top-3 similarities for the word \heb{הסתכלה}
(\textit{[she] looked [at]}), which occurs 17 times in the corpus, under the
different models.
Unsurprisingly, the lemma component has a positive effect on
semantics, while the tag component improves the morphological performance. It
also shows a clear trade-off between the two aspects -- as models which
perform the best on semantics are the worst on morphology. This behavior is
representative of the dozens of words we examined.

\paragraph{Quantitative Results}
We compare the different models on the different measures, and also compare to
the state-of-the-art n-gram based fastText model of Bojanowski et
al~\shortcite{bojanowski2016enriching} that does not require morphological
analysis. The results (Table~\ref{quantitative}) highlight the following:
\smallskip \newline
1. There is a trade-off between semantic and morphological performance -- improving one aspect comes at the expense of the other: the lemma component improves semantics but hurts morphology, while the opposite is true for the tag component. The common practice of using both components together is a kind of compromise: the \textit{LM}, \textit{WLM} and \textit{n-grams} models are not the best nor the worst on any measure.
\smallskip \newline
2. The impacts of the lemma and the tag components are much larger when dealing with rare words: comparing to \textit{W}, \textit{WL} is only 1.7\% better on \textit{SS} and 3.8\% worse on \textit{MS}, while it's 16.3\% better and 11.9\% worse on \textit{SSR} and \textit{MSR} (respectively). Similarly, \textit{WM} is only 2.8\% worse than \textit{W} on \textit{SS} and 44.9\% better on \textit{MS}, while it's 21.8\% worse and 75.7\% better on \textit{SSR} and \textit{MSR} (respectively).
\smallskip \newline
3. Simply lemmatizing the words is very effective for capturing semantic
similarity. This is especially true for the rare words, in which the L model
clearly outperform all others. For the common words, we see a small drop
compared to including the surface form as well (\textit{WL}, \textit{WLM}). This
is attributed to cases in which some of the semantics lies within the word's
morphological template, for example: in \textit{W} model, most similar words for
the masculine verb \heb{נפל} (\emph{fell}) are associated with \textit{a
soldier} (which is a masculine noun): \heb{נהרג} (was killed), \heb{נפגע} (was
injured), while the similarities of the feminine form \heb{נפלה} are associated
with \textit{a land} or \textit{a state} (both are feminine nouns): \heb{סופחה}
(was annexed), \heb{נכבשה} (was occupied). In \textit{L} model -- \heb{נפלה} and
\heb{נפל} share a single, less accurate representation (somewhat similarly to
representations of ambiguous words). This suggests using different compositions
for common and rare words.

\begin{table}
\small
\centering
\begin{tabular}{c||c|c||c|c}
	
     & \textit{SS} & \textit{SSR} & \textit{MS} & \textit{MSR}\\\hline
    \textit{W} & 0.707 & 0.675 & 0.626 & 0.569\\\hline
    \textit{L} & 0.713 & \textcolor{correct}{\textbf{0.816}} & \textcolor{wrong}{\textbf{0.491}} & \textcolor{wrong}{\textbf{0.339}}\\\hline
    \textit{WL} & \textcolor{correct}{\textbf{0.719}} & 0.785 & 0.602 & 0.501\\\hline
    \textit{WM} & \textcolor{wrong}{\textbf{0.687}} & \textcolor{wrong}{\textbf{0.528}} & \textcolor{correct}{\textbf{0.907}} & \textcolor{correct}{\textbf{1}}\\\hline
    \textit{LM} &  0.707 & 0.693 & 0.887 & 0.996\\\hline
    \textit{WLM} & 0.716 & 0.748 & 0.882 & \textcolor{correct}{\textbf{1}}\\\hline
    \textit{n-grams} & 0.712 & 0.767 & 0.71 & 0.866\\\hline
\end{tabular}
\caption{\label{quantitative} \footnotesize Results on \textit{SemanticSim (SS)}, \textit{SemanticSimRare (SSR)}, \textit{MorphoSim (MS)} and \textit{MorphoSimRare (MSR)}. The best result for each measure is green, the worst is red.}
\end{table}

\section{Conclusions}
\vspace{-5pt}

Our key message is that users of morphology-driven models should consider the trade-off between the different components of their representations.
Since the goal of most works on morphology-driven models was to improve \textit{semantic} similarity, the configurations they used (which combine both semantic and morphological components) were probably not the best choices: we show that using the lemma component (either alone or together with the surface form) is better.
Indeed, excluding the morphological component will make the morphological similarity drop, but it's not necessarily a problem for every task. One should include the morphological component in the embeddings only for tasks in which morphological similarity is required and cannot be handled by other means. A future work can be to perform an extrinsic evaluation of the different models in various downstream applications. This may reveal which kinds of tasks benefit from morphological information, and which can be done better by a pure semantic model.

\section*{Acknowledgements}
The work was supported by the Israeli Science Foundation (grant number 1555/15).

\bibliography{eacl2017}
\bibliographystyle{eacl2017}

\end{document}